\documentclass[11pt,a4paper]{article}
\usepackage[hyperref]{acl2020}
\usepackage{times}
\usepackage{latexsym}

\usepackage{microtype}
\aclfinalcopy
\def\aclpaperid{2920}

\usepackage{ifthen}
\usepackage{amsmath}
\usepackage{amsthm}
\usepackage{amssymb}
\usepackage{graphicx}
\usepackage{subfigure}
\usepackage{comment}
\usepackage{multirow}
\usepackage{algorithm}
\usepackage{algorithmic}
\usepackage{xcolor}
\usepackage{xspace}

\newif\ifcomment
\commentfalse

\newcommand\sT{\ensuremath{\mathcal{T}}}

\newcommand\sX{\ensuremath{\mathcal{X}}}
\newcommand\sY{\ensuremath{\mathcal{Y}}}
\newcommand\sZ{\ensuremath{\mathcal{Z}}}

\newcommand\R{\ensuremath{\mathbb{R}}} 

\newcommand\refsec[1]{Section~\ref{sec:#1}}

\newcommand\reffig[1]{Figure~\ref{fig:#1}}

\newcommand\reftab[1]{Table~\ref{tab:#1}}

\ifthenelse{\isundefined{\definition}}{\newtheorem{definition}{Definition}}{}
\ifthenelse{\isundefined{\assumption}}{\newtheorem{assumption}{Assumption}}{}
\ifthenelse{\isundefined{\hypothesis}}{\newtheorem{hypothesis}{Hypothesis}}{}
\ifthenelse{\isundefined{\proposition}}{\newtheorem{proposition}{Proposition}}{}
\ifthenelse{\isundefined{\theorem}}{\newtheorem{theorem}{Theorem}}{}
\ifthenelse{\isundefined{\lemma}}{\newtheorem{lemma}{Lemma}}{}
\ifthenelse{\isundefined{\corollary}}{\newtheorem{corollary}{Corollary}}{}
\ifthenelse{\isundefined{\alg}}{\newtheorem{alg}{Algorithm}}{}
\ifthenelse{\isundefined{\example}}{\newtheorem{example}{Example}}{}

\ifcomment
\newcommand\pl[1]{\textcolor{red}{[PL: #1]}}
\newcommand{\ar}[1]{{\color{blue} AR: #1}}
\newcommand{\rj}[1]{{\bf \color{violet} [RJ: #1]}}
\newcommand{\ej}[1]{{\bf \color{brown} [EJ: #1]}}
\else
\newcommand\pl[1]{}
\newcommand{\ar}[1]{}
\newcommand{\rj}[1]{}
\newcommand{\ej}[1]{}
\fi

\newcommand{\stdacc}{\text{acc}_\text{std}}
\newcommand{\robacc}{\text{acc}_\text{rob}}
\newcommand{\atkacc}{\text{acc}_\text{attack}}

\newcommand{\ftask}{f}

\newcommand{\falpha}{f_\alpha}
\newcommand{\fenc}{g}

\newcommand{\xt}{\tilde{x}}

\newcommand{\phirep}{\alpha}
\newcommand{\phitok}{\pi}
\newcommand{\phitokbar}{\pi_V}
\newcommand{\phitokoov}{\pi_{\text{OOV}}}

\newcommand{\Brep}{B_{\alpha}}
\newcommand{\Btok}{B_{\pi}}
\newcommand{\Ztok}{\sZ_{\text{Tok}}}
\newcommand{\tokens}{\sT}
\newcommand{\oov}{\text{OOV}}
\newcommand{\nl}[1]{{\it ``#1''}}
\newcommand{\conncomp}{\textsc{ConnComp}\xspace} 
\newcommand{\agglom}{\textsc{AggClust}\xspace}
\newcommand{\concomptext}{\text{Con. Comp. }}
\newcommand{\agglomtext}{\text{Agg. Clust. }}

\newcommand{\ptask}{p_\text{task}}
\newcommand{\Pcorp}{P_\mathsf{x}}
\newcommand{\Stability}{\operatorname{Stab}}
\newcommand{\Fidelity}{\operatorname{Fid}}

\title{Robust Encodings: A Framework for Combating Adversarial Typos}

\author{Erik Jones\hspace{6mm}Robin Jia\thanks{\ Authors contributed equally.}\hspace{6mm}Aditi Raghunathan\footnotemark[1]\hspace{6mm}Percy Liang\\
  Computer Science Department, Stanford University \\
  \texttt{\{erjones,robinjia,aditir,pliang\}@stanford.edu} \\}

\date{}

\begin{document}
\maketitle
\begin{abstract}
Despite excellent performance on many tasks, NLP systems are easily fooled by small adversarial perturbations of inputs. Existing procedures to defend against such perturbations are either (i) heuristic in nature and susceptible to stronger attacks or 
(ii) provide guaranteed robustness to worst-case attacks, but are incompatible with state-of-the-art models like BERT.
In this work, we introduce \emph{robust encodings} (RobEn): a simple framework that confers guaranteed robustness, without making compromises on model architecture. 
The core component of RobEn is an \emph{encoding function}, which maps sentences to a smaller, 
discrete space of encodings.
Systems using these encodings as a bottleneck confer guaranteed robustness with \emph{standard training}, and the same encodings can be used across multiple tasks.
We identify two desiderata to construct robust encoding functions:
perturbations of a sentence should map to a small set of encodings 
(stability), and models using encodings should still perform well (fidelity).
We instantiate RobEn to defend against a large family of adversarial typos. 
Across six tasks from GLUE,
our instantiation of RobEn paired with BERT achieves an average robust accuracy of $71.3\%$ against \emph{all} adversarial typos in the family considered, while previous work using a typo-corrector achieves only $35.3\%$ accuracy against a simple greedy attack.

\end{abstract}

\section{Introduction}
State-of-the-art NLP systems are brittle: small perturbations of inputs, commonly referred to as adversarial examples, can lead to catastrophic model failures \citep{belinkov2018synthetic,ebrahimi2018hotflip,ribeiro2018sears,alzantot2018adversarial}. 
For example, carefully chosen typos and word substitutions have fooled systems for hate speech detection \citep{hosseini2017deceiving}, machine translation \citep{ebrahimi2018adversarial},
and spam filtering \citep{lee2005spam}, among others.

\begin{figure}
    \centering
    \includegraphics[width=0.95\linewidth]{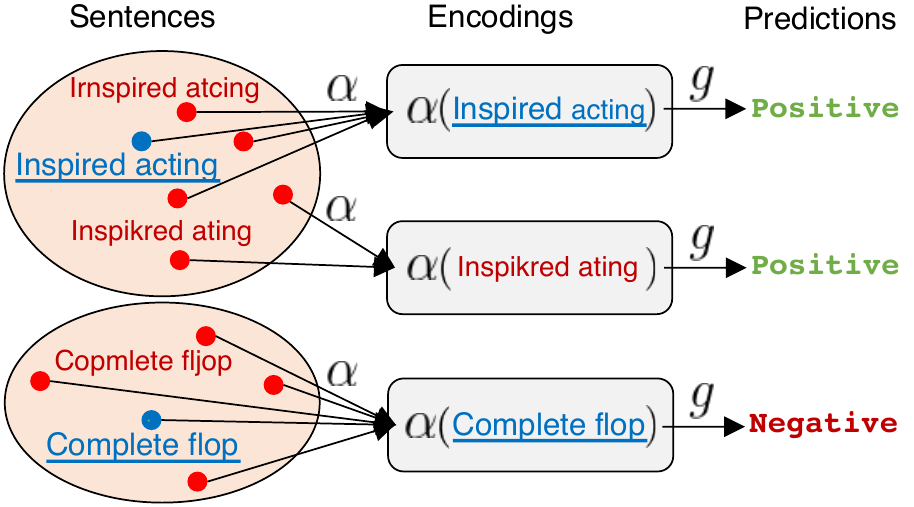}
    \caption{Example of a defense using RobEn. An adversary can perturb sentences (blue, underlined) to many different perturbations (red, not-underlined) within the attack surface (red, ovals). We define an encoding function $\phirep$ such that each perturbation of the input sentences maps to one of a few encodings (grey, rounded rectangles). We can then use any model $\fenc$ to make predictions given the encodings.
    }
    \label{fig:roben}
\end{figure}

We aim to build systems that achieve high \emph{robust accuracy}: accuracy against worst-case attacks. Broadly, existing methods to build robust models fall under one of two categories: (i) adversarial training, which augments the training set with heuristically generated perturbations and 
(ii) certifiably robust training, which bounds the change in prediction between an input and \emph{any} of its allowable perturbations. 
Both these approaches have major shortcomings, especially in NLP.
Adversarial training, while quite successful in vision~\cite{madry2018towards}, is challenging in NLP  
due to the discrete nature of textual inputs 
\citep{ebrahimi2018hotflip}; current techniques like projected gradient descent are incompatible with subword tokenization. Further, adversarial training relies on heuristic approximations to the worst-case perturbations, 
leaving models vulnerable to new, stronger attacks.
Certifiably robust training \citep{jia2019certified, huang2019achieving, shi2020robustness} circumvents the above challenges by optimizing over a convex outer-approximation of the set of perturbations, allowing us to lower bound the true robust accuracy.
However, the quality of bounds obtained by these methods scale poorly with the size of the network, and are vacuous for state-of-the-art models like BERT. 
Moreover, both approaches require separate, expensive training for each task, even when defending against the same type of perturbations. 

Ideally we would like a ``robustness'' module that we can reuse across multiple tasks, allowing us to only worry about robustness once: during its construction. Indeed, \emph{reusable} components have driven recent progress in NLP.
For example, word vectors are a universal resource that are constructed once, then used for many different tasks. 
Can we build a reusable robust defense that can easily work with complex, state-of-the-art architectures like BERT?
The recent work of \citet{pruthi2019misspellings}, which uses a typo-corrector to defend against adversarial typos, is such a reusable defense: it is trained once, then reused across different tasks.
However, we find that current typo-correctors do not perform well against even heuristic attacks, limiting their applicability. 

Our primary contribution is \emph{robust encodings} (RobEn), a framework to construct encodings that can make systems using \emph{any} model robust.
The core component of RobEn is an \emph{encoding function} that maps sentences to a smaller discrete space of encodings, which are then used to make predictions. 
We define two desiderata that a robust encoding function should satisfy: stability and fidelity. 
First, to encourage consistent predictions across perturbations, 
the encoding function should map all perturbations of a sentence to a small set of encodings (stability). 
Simultaneously, encodings should remain expressive, so models trained using encodings still perform well on unperturbed inputs (fidelity).
Because systems using RobEn are encoding-based we can compute the \emph{exact} robust accuracy tractably, avoiding the lower bounds of certifiably robust training.
Moreover, these encodings can make any downstream model robust, including state-of-the-art transformers like BERT, and can be reused across different tasks.

In Section \ref{sec:roben_cluster}, we apply RobEn to combat adversarial typos.
In particular, we allow an attacker to add independent edit distance one typos to each word in an input sentence, resulting in exponentially more possible perturbations than previous work \citep{pruthi2019misspellings, huang2019achieving}.  
We consider a natural class of \emph{token-level encodings}, which are obtained by encoding each token in a sentence independently. This structure allows us to express stability and fidelity in terms of a clustering objective, which we optimize.

Empirically, our instantiation of RobEn 
achieves state-of-the-art robust accuracy, which we compute exactly, across six classification tasks from the GLUE benchmark \citep{wang2019glue}. 
Our best system, which combines RobEn with a BERT classifier \citep{devlin2019bert},
achieves an average robust accuracy of $71.3\%$ across the six tasks.
In contrast, 
a state-of-the-art defense that combines BERT with a typo corrector \citep{pruthi2019misspellings} gets $35.3\%$ accuracy when adversarial typos are inserted,
and a standard data augmentation defense gets only $12.2\%$ accuracy.

\section{Setup}
\label{sec:setup}
\paragraph{Tasks.} 
We consider NLP tasks that require classifying textual input $x \in \sX$ to a class $y \in \sY$.
For simplicity, we refer to inputs as sentences.
Each sentence $x$ consists of tokens $x_1, \dotsc, x_L$ from the set of all strings $\sT$.
Let $\ptask$ denote the distribution over inputs and labels for a particular task of interest. 
The goal is to learn a model $\ftask: \sX \rightarrow \sY$ that maps sentences to labels, given training examples $(x, y) \sim \ptask$.

\paragraph{Attack surface.} 
We consider an attack surface in which an adversary can perturb each token $x_i$ of a sentence to some token $\tilde{x_i} \in B(x_i)$, where $B(x_i)$ is the set of valid perturbations of $x_i$. 
\begin{figure}
    \centering
    \includegraphics[width=0.95\linewidth]{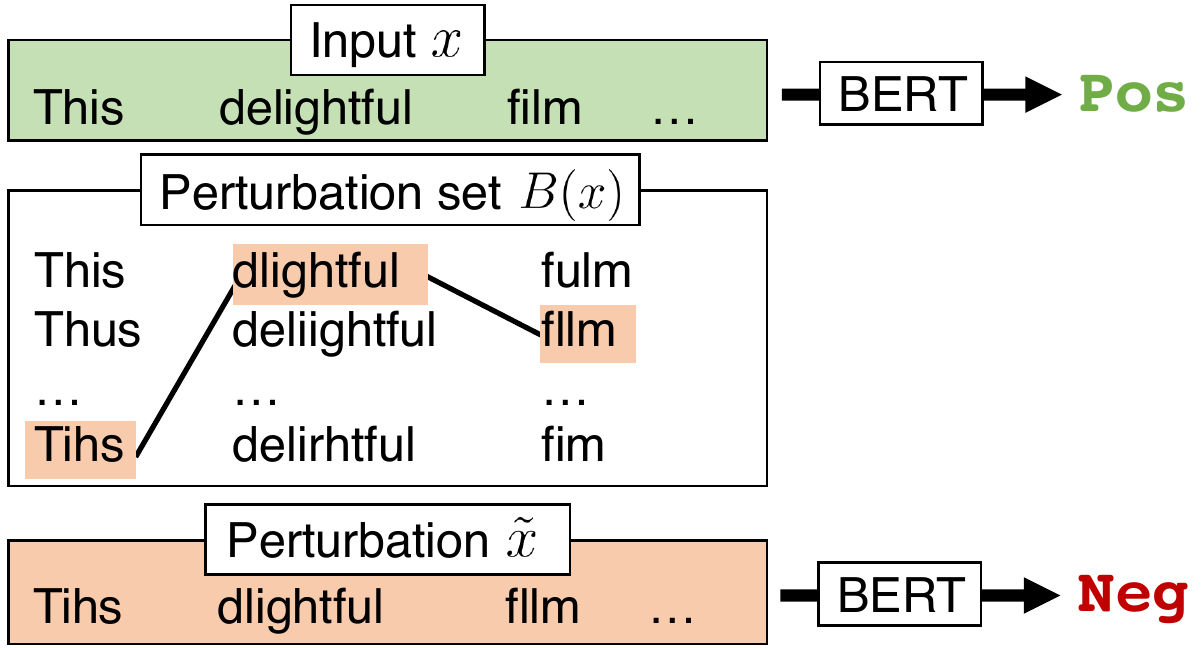}
    \caption{Attack model allowing independent perturbations of each token. The original input, $x$ is classified by the model as positive while the perturbation $\tilde{x} = $, obtained by choosing perturbations of \nl{This}, \nl{delightful}, and \nl{film} independently, is classified as negative. Independent perturbations of each word results in an exponentially large perturbation space $B(x)$.
    }
    \label{fig:attack}
\end{figure}
For example, $B(x_i)$ could be a set of allowed typos of $x_i$.
We define $B(x)$ as the set of all valid perturbations of the set $x$, where every possible combination of token-level typos is allowed:
\begin{align}
B(x) = \{(\tilde{x_1}, \dotsc, \tilde{x_L}) \mid \tilde{x_i} \in B(x_i) \, \forall \, i\}
\end{align}
The size of the attack surface $|B(x)|$ grows exponentially with respect to number of input tokens, as shown in \reffig{attack}. In general $x_i \in B(x_i)$, so some words could remain unperturbed. 

\paragraph{Model evaluation.}
In this work, we use three evaluation metrics for any given task.

First, we evaluate a model on its \emph{standard accuracy} on the task: 
\begin{align}
    \stdacc(\ftask) &= \mathbb{E}_{(x, y) \sim \ptask}\mathbf{1}[\ftask(x) = y].
\end{align}

Next, we are interested in models that also have high \emph{robust accuracy}, the fraction of examples $(x, y)$ for which the model is correct on all valid perturbations $\xt \in B(x)$ allowed in the attack model:
\begin{align}
    \robacc(\ftask) &= \mathbb{E}_{(x, y) \sim \ptask}\min_{\xt \in B(x)}\mathbf{1}\left[\ftask(\xt) = y\right].
\end{align}
It is common to instead compute accuracy against a heuristic attack $a$ that maps clean sentences $x$ to perturbed sentences $a(x) \in B(x)$.
\begin{align}
    \atkacc(\ftask; a) &= \mathbb{E}_{(x, y) \sim \ptask}\mathbf{1}[\ftask(a(x)) = y].
\end{align}
 Typically, $a(x)$ is the result of a heuristic search for a perturbation $\xt \in B(x)$ that $\ftask$ 
misclassifies.
Note that $\atkacc$ is a (possibly loose) upper bound of $\robacc$ because there could be perturbations that the model misclassifies but are not encountered during the heuristic search \citep{athalye2018obfuscated}. 

Additionally, since robust accuracy is generally hard to compute, some existing work computes \emph{certified accuracy} \citep{huang2019achieving, jia2019certified, shi2020robustness}, which is a potentially conservative lower bound for the true robust accuracy. In this work, since we use robust encodings, we can tractably compute the exact robust accuracy. 
\section{Robust Encodings}
We introduce \emph{robust encodings} (RobEn), a framework for constructing encodings that are reusable across many tasks, and pair with arbitrary model architectures. In \refsec{roben_general} we describe the key components of RobEn, then in \refsec{encoding_function_desiderata} we highlight desiderata RobEn should satisfy.
\subsection{Encoding functions}
\label{sec:roben_general}
A RobEn classifier $\falpha: \sX \rightarrow \sY$ using RobEn decomposes into two components: a \emph{fixed} encoding function $\phirep: \sX \rightarrow \sZ$, and a model that accepts encodings $\fenc: \sZ \rightarrow \sY$.\footnote{We can set $\sZ \subseteq \sX$ when $\fenc$ accepts sentences.} For any sentence $x$, our system makes the prediction $\falpha(x) = \fenc(\phirep(x))$. 
Given training data 
$\{(x_i, y_i)\}_{i=1}^n$ and the encoding function $\phirep$, we learn $\fenc$ by performing standard training on encoded training points $\{(\phirep(x_i), y_i)\}_{i=1}^n$. To compute the robust accuracy of this system, we note that for well-chosen $\phirep$ and an input $x$ from some distribution $\Pcorp$, the set of possible encodings $\alpha(\tilde{x})$ for some perturbation $\tilde{x} \in B(x)$ is both small and tractable to compute quickly. We can thus compute $\robacc(\falpha)$ quickly by generating this set of possible encodings, and feeding each into $\fenc$, which can be any architecture.

\subsection{Encoding function desiderata}
\label{sec:encoding_function_desiderata}
In order to achieve high robust accuracy, a classifier $\falpha$ that uses $\phirep$ should make consistent predictions on all $\xt \in B(x)$, the set of points described by the attack surface, and also have high standard accuracy on unperturbed inputs.
We term the former property \emph{stability}, and the latter \emph{fidelity}, give intuition for both in this section, and provide a formal instantiation in Section \ref{sec:roben_cluster}. 

\paragraph{Stability.}
For an encoding function $\phirep$ and some distribution over inputs $\Pcorp$, the stability $\Stability(\phirep)$ measures how often $\phirep$ maps sentences $x \sim \Pcorp$ to the same encoding as all of their perturbations.

\paragraph{Fidelity.}
An encoding function $\phirep$ has high fidelity if models that use $\phirep$ can still achieve high standard accuracy. Unfortunately, while we want to make task agnostic encoding functions, standard accuracy is inherently \emph{task dependent}: different tasks have different expected distributions over inputs and labels. To emphasize this challenge consider two tasks: for an integer $n$, predict $n$ mod $2$, and $n$ mod $3$. The information we need encodings to preserve varies significantly between these tasks: for the former, $2$ and $6$ can be identically encoded, while for the latter they must encoded separately.

To overcome this challenge, we consider a single distribution over the inputs $\Pcorp$ that we believe covers many task-distributions $\ptask$. Since it is hard to model the distribution over the labels, we take the more conservative approach of mapping the different sentences sampled from $\Pcorp$ to different encodings with high probability. We call this $\Fidelity(\phirep)$, and give an example in \refsec{agg_clust}.

\paragraph{Tradeoff.}
Stability and fidelity are inherently competing goals. An encoding function that maps every sentence to the same encoding trivially maximizes stability, but is useless for any non-trivial classification task. Conversely, fidelity is maximized when every input is mapped to itself, which has very low stability. 
In the following section, we construct an instantiation of RobEn that balances stability and fidelity when the attack surface consists of typos. 
\section{Robust Encodings for Typos}
\label{sec:roben_cluster}
In this section, we focus on \emph{adversarial typos}, where an adversary can add typos to each token in a sentence (see \reffig{attack}). 
Since this attack surface is defined at the level of tokens, we restrict attention to encoding functions that encode each token independently. Such an encoding does not use contextual information; we find that even such robust encodings achieve greater attack accuracy and robust accuracy in practice than previous work.

First, we will reduce the problem of generating token level encodings to assigning vocabulary words to clusters (Section \ref{sec:reduction}). Next, we use an example to motivate different clustering approaches (Section \ref{sec:tokenexample}), then describe how we handle out-of-vocabulary tokens (Section \ref{sec:oov}). Finally, we introduce two types of token-level robust encodings: connected component encodings (Section \ref{sec:conncomp}) and agglomerative cluster encodings (Section \ref{sec:agg_clust}).

\subsection{Encodings as clusters}
\label{sec:reduction}
We construct an encoding function $\alpha$ that encodes $x$ token-wise.
Formally, $\alpha$ is defined by a token-level encoding function $\pi$ that
maps each token $x_i \in \sT$ to some \emph{encoded token} $\phitok(x_i) \in \Ztok$:
\begin{align}
\alpha(x) = [\phitok(x_1), \phitok(x_2), \hdots \phitok(x_L)].
\end{align}
In the RobEn pipeline, a downstream model $g$ is trained on encodings (Section~\ref{sec:roben_general}). 
If $\phitok$ maps many words and their typos to the same encoded token, they become indistinguishable to $g$, conferring robustness. 
In principle, the relationship between different encoded tokens is irrelevant: during training, $g$ learns how to use the encoded tokens to perform a desired task. 
Thus, the problem of finding a good $\phitok$ is equivalent to deciding which tokens should share the same encoded token.

Since the space of possible tokens $\sT$ is innumerable, we focus on a smaller set of words $V = \{w_1, \hdots, w_N\} \subseteq \tokens$, which contains the $N$ most frequent words over 
$\Pcorp$. We will call elements of $V$ \emph{words}, and tokens that are perturbations of some word \emph{typos}. We view deciding which words should share an encoded token as assigning words to clusters $C_1, \hdots, C_k \subseteq V$. For all other tokens not in the vocabulary, including typos, we define a separate $\phitokoov$. Thus, we decompose $\phitok$ as follows: 
\begin{align}
    \phitok(x_i) &= \begin{cases}\phitokbar(x_i) & x_i \in V \\ \phitokoov(x_i) & x_i \notin V \end{cases},
\end{align}
Here, $\phitokbar$ is associated with a clustering $C$ of vocabulary words, where each cluster is associated with a unique encoded token.
\begin{figure}
\center
\includegraphics[width = 0.95\linewidth]{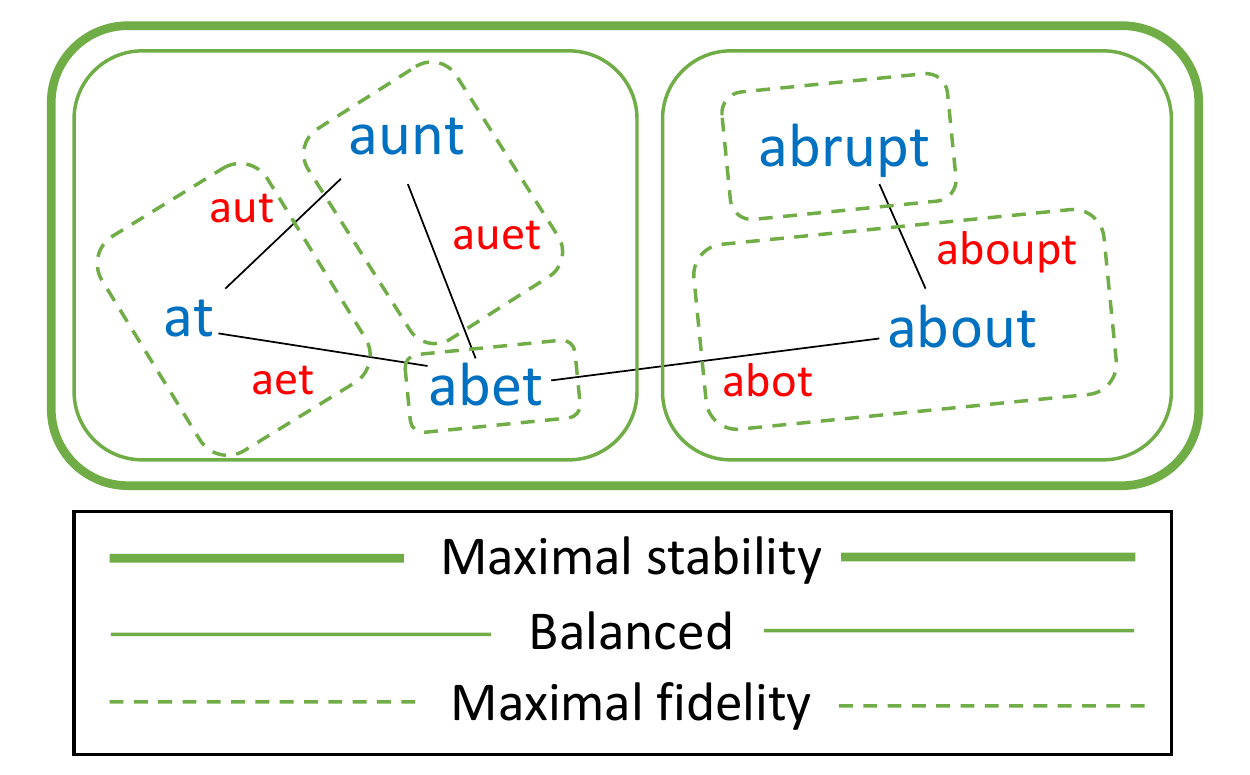}
\caption{Visualization of three different encodings.
Vocabulary words (large font, blue) share an edge if they share a common perturbation (small font, red). The maximal stability cluster (thick solid line) clusters identically, the maximal fidelity clusters (thin dotted line) encodes all words separately, while the balanced clusters (thin solid line) trade off the two.
} 
\label{fig:cluster_graph}
\end{figure}

\subsection{Simple example}
\label{sec:tokenexample}
We use a simple example to illustrate how a token-level encoding function can achieve the RobEn desiderata: stability and fidelity defined in Section \ref{sec:encoding_function_desiderata}. 
We will formally define the stability and fidelity of a clustering in Sections~\ref{sec:oov} and \ref{sec:agg_clust}.

Consider the five words (large font, blue) in \reffig{cluster_graph}, 
along with potential typos (small font, red).
We illustrate three different clusterings as boxes around tokens in the same cluster. 
We may put all words in the same cluster (thick box), each word in its own cluster (dashed boxes), or something in between (thin solid boxes). 
For now, we group each typo with a word it could have been perturbed from (we will discuss this further in \refsec{oov}).

To maximize stability, we need to place all words in the same cluster. Otherwise, there would be two words (say \nl{at} and \nl{aunt}) that could both be perturbed to the same typo (\nl{aut}) but are in different clusters. Therefore, \nl{aut} cannot map to the same encoded token as both the possible vocab words. At the other extreme, to maximize fidelity, each word should be in its own cluster.
Both mappings have weaknesses: the stability-maximizing mapping has low fidelity since all words are identically encoded and thus indistinguishable, while the fidelity-maximizing mapping has low stability since the typos of words \nl{aunt}, \nl{abet}, and \nl{abrupt} could all be mapped to different encoded tokens than that of the original word. 

The clustering represented by the thin solid boxes in \reffig{cluster_graph} balances stability and fidelity. Compared to encoding all words identically, it has higher fidelity, since it distinguishes between some of the words (e.g., \nl{at} and \nl{about} are encoded differently). It also has reasonably high stability, since only the infrequent \nl{abet} has typos that are shared across words and hence are mapped to different encoded tokens. 
\subsection{Encoding out-of-vocab tokens} 
\label{sec:oov}
Given a fixed clustering of $V$, we now study how to map out-of-vocabulary tokens, including typos, to encoded tokens without compromising stability.

\paragraph{Stability.} Stability measures the extent to which typos of words map to different encoded tokens. We formalize this by defining the set of tokens that some typo of a word $w$ could map to, $\Btok(w)$:
\begin{align}
    \Btok(w) = \{ \phitok(\tilde{w}); \tilde{w} \in B(w) \}, 
\end{align}
where $B(w)$ is the set of allowable typos of $w$. 
Since we care about inputs drawn from $\Pcorp$, we define $\Stability$ on the clustering $C$ using $\rho(w)$, the normalized frequency of word $w$ based on $\Pcorp$.
\begin{align}
    \Stability(C) &= -\sum_{i=1}^N \rho(w_i) |\Btok(w_i)|
    \label{eqn:stability}
\end{align}
For a fixed clustering, the size of $\Btok(w)$ depends on where $\phitokoov$ maps typos that $w$ shares with other words; for example in \reffig{cluster_graph}, \nl{aet} could be a perturbation of both \nl{at} and \nl{abet}. If we map the typo the encoded token of \nl{at}, we increase the size of $\Btok(\text{"abet"})$ and vice-versa. In order to keep the size of $\Btok(w)$ smaller for the more frequent words and maximize stability (Equation~\ref{eqn:stability}), we map a typo to the same encoded token as its most frequent neighbor word (in this case \nl{at}). 
Finally, when a token is not a typo of any vocab words, we encode it to a special token $\oov$. 
\subsection{Connected component encodings}
\label{sec:conncomp}
We present two approaches to generate robust token-level encodings. Our first method, connected component encodings, maximizes the stability objective (\ref{eqn:stability}). Notice that $\Stability$ is maximized when for each word $w$, $\Btok(w)$ contains one encoded token. This is possible only when all words that share a typo are assigned to the same cluster. 

To maximize $\Stability$, define a graph $G$ with all words in $V$ as vertices, and edges between words that share a typo. Since we must map words that share an edge in $G$ to the same cluster, we define the cluster $C_i$ to be the set of words in the $i^{th}$ connected component of $G$. 
While this stability-maximizing clustering encodes many words to the same token (and hence seems to compromise on fidelity), these encodings still perform surprisingly well in practice (see \refsec{roben_results}).

\subsection{Agglomerative cluster encodings}
\label{sec:agg_clust}
Connected component encodings focus only stability and can lead to needlessly low fidelity. For example, in \reffig{cluster_graph}, \nl{at} and \nl{about} are in the same connected component even though they don't share a typo.  
Since both words are generally frequent, mapping them to different encoded tokens can significantly improve fidelity, with only a small drop in stability: recall only the infrequent word \nl{abet} can be perturbed to multiple encoded tokens.

To handle such cases, we introduce \emph{agglomerative cluster encodings}, which we construct by trading off $\Stability$ with a formal objective we define for fidelity: $\Fidelity$.
We then approximately optimize this combined objective $\Phi$ using an agglomerative clustering algorithm.

\paragraph{Fidelity objective.} 
Recall from Section \ref{sec:encoding_function_desiderata} that an encoding has high fidelity if it can be used to achieve high standard accuracy on many tasks. This is hard to precisely characterize: we aim to design an objective that could approximate this. 

We note that distinct encoded tokens are arbitrarily related: the model $g$ learns how to use different encodings during training. Returning to our example, suppose \nl{at} and \nl{abet} belong to the same cluster and share an encoded token $z$. During training, each occurrence of \nl{at} and \nl{abet} is replaced with $z$. However, since \nl{at} is much more frequent, classifiers treat $z$ similarly to $\nl{at}$ in order to achieve good overall performance.
This leads to mostly uncompromised performance on sentences with \nl{at}, at the cost of performance on sentences containing the less frequent \nl{abet}. 

This motivates the following definition: let $\vec{v}_i$ be a the indicator vector in $\R^{|V|}$ corresponding to word $i$. In principle $\vec{v}_i$ could be a word embedding; we choose indicator vectors to avoid making additional assumptions. We define the encoded token $\vec{\mu_j}$ associated with words in cluster $C_j$ as follows:
\begin{align}
    \vec{\mu_j} &= \frac{\sum_{w_i \in C_j} \rho(w_i) \vec{v_i}}{\sum_{w_i \in C_j}  \rho(w_i)}
\end{align}
We weight by the frequency $\rho$ to capture the effect of training on the encodings, as described above. 

Fidelity is maximized when each word has a distinct encoded token. We capture the drop in standard accuracy due to shared encoded tokens by computing the distance between the original embeddings of the word its encoded token. Formally, let $c(i)$ be the cluster index of word $w_i$. We define the fidelity objective $\Fidelity$ as follows:
\begin{align}
    \Fidelity(C) &= -\sum_{i=1}^N \rho(w_i) \| \vec{v_i} - \vec{\mu}_{c(i)}\|^2.
    \label{eqn:fidelity}
\end{align}
$\Fidelity$ is high if frequent words and rare words are in the same cluster and is low when when multiple frequent words are in the same cluster.

\paragraph{Final objective.} 
We introduce a hyperparameter $\gamma \in [0,1]$ that balances stability and fidelity. We approximately minimize the following weighted combination of $\Stability$ (\ref{eqn:stability}) and $\Fidelity$ (\ref{eqn:fidelity}):
\begin{align}
    \Phi(C) &= \gamma \Fidelity(C) + (1 - \gamma) \Stability(C).
\end{align}
As $\gamma$ approaches 0, we get the connected component clusters from our baseline, which maximize stability. As $\gamma$ approaches 1, we maximize fidelity by assigning each word to its own cluster.

\paragraph{Agglomerative clustering.}
 We approximate the optimal value of $\Phi$ using \emph{agglomerative clustering}; we start with each word in its own cluster, then iteratively combine the pair of clusters whose resulting combination increases $\Phi$ the most. We repeat until combining any pair of clusters would decrease $\Phi$. Further details are provided in Appendix \ref{apdx:aggalgo}.
\section{Experiments}
\subsection{Setup}
\label{sec:experiments_setup}
\paragraph{Token-level attacks.} The primary attack surface we study is edit distance one (ED1) perturbations. For every word in the input, the adversary is allowed to insert a lowercase letter, delete a character, substitute a character for any letter, or swap two adjacent characters, so long as the first and last characters remain the same as in the original token. The constraint on the outer characters, also used by \citet{pruthi2019misspellings}, is motivated by psycholinguistic studies \citep{rawlison1976letterpos, davis2003psycholinguistic}.

Within our attack surface, \nl{the movie was miserable} can be perturbed to \nl{thae mvie wjs misreable}  but not \nl{th movie as miserable}. Since each token can be independently perturbed, the number of perturbations of a sentence grows exponentially with its length; even \nl{the movie was miserable} has 431,842,320 possible perturbations. Our attack surface contains
the attack surface used by \cite{pruthi2019misspellings}, which allows ED1 perturbations to at most two words per sentence. Reviews from SST-2 have 5 million perturbations per example (PPE) on average under this attack surface, while our attack surface averages $10^{97}$ PPE. 
We view the size of the attack surface as a strength of our approach: our attack surface forces a system robust to subtle perturbations (\nl{the moviie waas misreable}) that smaller attack surfaces miss.

In Section \ref{sec:intprm}, we additionally consider the internal permutation attacks studied in \citet{belinkov2018synthetic} and \citet{sakaguchi2017robsut}, where all characters, except the first and the last, may be arbitrarily reordered.

\paragraph{Attack algorithms.}
We consider two attack algorithms: the worst-case attack (WCA) and a beam-search attack (BSA). 
WCA exhaustively tests every possible perturbation of an input $x$ to see any change in the prediction. 
The attack accuracy of WCA is the true robust accuracy since if there exists some perturbation that changes the prediction, WCA finds it. 
When instances of RobEn have high stability, the number of possible encodings of perturbations of $x$ is often small, allowing us to exhaustively test all possible perturbations in the encoding space.\footnote{When there are more than 10000 possible encodings, which holds for $0.009\%$ of our test examples, we assume the adversary successfully alters the prediction.} 
This allows us to tractably run WCA. Using WCA with RobEn, we can obtain computationally tractable \emph{guarantees} on robustness: given a sentence, we can quickly compute whether or not any perturbation of $x$ that changes the prediction.

For systems that don't use RobEn, we cannot tractably run WCA. Instead, we run a beam search attack (BSA) with beam width 5, perturbing tokens one at a time.
For efficiency, we sample at most $\operatorname{len}(x_i)$ perturbations at each step of the search (see Apendix \ref{apdx:attacks}). 
Even against this very limited attack, we find that baseline models have low accuracy.

\paragraph{Datasets.}
We use six of the nine tasks from GLUE \citep{wang2019glue}: SST-2, MRPC, QQP, MNLI, QNLI, and RTE. 
We do not use STS-B and CoLA as they are evaluated on correlation, which does not decompose as an example-level loss.
We additionally do not use WNLI, as most submitted GLUE models cannot even outperform the majority baseline, and state-of-the-art models are rely on external training data \citep{kocijan2019winograd}.
We evaluate on the test sets for SST-2 and MRPC, and the publicly available dev sets for the remaining tasks.
More details are provided in Appendix~\ref{sec:datasets}.

\subsection{Baseline models.}
We consider three baseline systems. Our first is the standard base uncased BERT model \citep{devlin2019bert} fine-tuned on the training data for each task.\footnote{\url{https://github.com/huggingface/pytorch-transformers}}

\paragraph{Data augmentation.} For our next baseline, we augment the training dataset with four random perturbations of each example, then fine-tune BERT on this augmented data.
Data augmentation has been shown to increase robustness to some types of adversarial perturbations \citep{ribeiro2018sears, liu2019inoculation}. Other natural baselines all have severe limitations. Adversarial training with black-box attacks offers limited robustness gains over data augmentation \citep{cohen2019certified, pruthi2019misspellings}. Projected gradient descent \citep{madry2017towards}, the only white-box adversarial training method that is robust in practice, cannot currently be applied to BERT since subword tokenization maps different perturbations to different numbers of tokens, making gradient-based search impossible. 
Certifiably robust training \citep{huang2019achieving, shi2020robustness} does not work with BERT due to the same tokenization issue and BERT's use of non-monotonic activation functions, which make computing bounds intractable. Moreover the bounds computed with certifiably robust training, which give guarantees, become loose as model depth increases, hurting robust performance \citep{gowal2018effectiveness}.

\paragraph{Typo-corrector.} For our third baseline, we use the most robust method from \citet{pruthi2019misspellings}. In particular, we train a scRNN typo-corrector \citep{sakaguchi2017robsut} on random perturbations of each task's training set. At test time inputs are ``corrected'' using the typo corrector, then fed into a downstream model.
We replace any $\oov$ outputted by the typo-corrector with the neutral word \nl{a} and use BERT as our downstream model.

\subsection{Models with RobEn}
We run experiments using our two token-level encodings: connected component encodings ($\conncomp$) and agglomerative cluster encodings ($\agglom$). 
To form clusters, we use the $N=100,000$ most frequent words from the Corpus of Contemporary American English \citep{davies2008COCA} that are also in GloVe \citep{pennington2014glove}.
For $\agglom$ we use $\gamma = 0.3$, which maximizes robust accuracy on SST-2 dev set. 

\paragraph{Form of encodings.} Though unnecessary when training from scratch, to leverage the inductive biases of pre-trained models like BERT \citep{devlin2019bert}, we define the encoded token of a cluster to be the cluster's most frequent member word. 
In the special case of the out-of-vocab token, we map $\oov$ to $[\text{MASK}]$. 
Our final encoding, $\phirep(x)$, is the concatenation of all of these words.
For both encodings, we fine-tune BERT on the training data, using $\phirep(x)$ as input.
Further details are in Appendix~\ref{apdx:hparams}.

\subsection{Robustness gains from RobEn}
\label{sec:roben_results}
\begin{table*}[t]
  \small
  \centering
  \begin{tabular}{|l|l|rrrrrrr|}
    \hline
    Accuracy & \multicolumn{1}{c|}{System} 
    & \multicolumn{1}{c}{SST-2}
    & \multicolumn{1}{c}{MRPC}
    & \multicolumn{1}{c}{QQP}
    & \multicolumn{1}{c}{MNLI}
    & \multicolumn{1}{c}{QNLI}
    & \multicolumn{1}{c}{RTE}
    & \multicolumn{1}{c|}{Avg} \\ \hline
    &&&&&&&&\\[-2.5mm]
    \multirow{7}{*}{Standard} & \textbf{Baselines} & & & & & & & \\
    & BERT & $93.8$ & $87.7$ & $91.3$ & $84.6$ & $88.6$ & $71.1$ &  $86.2$ \\
    & Data Aug. + BERT & $92.2$ & $84.3$ & $88.7$ & $83.0$ & $87.4$ & $63.5$ & $83.1$ \\
    & Typo Corr. + BERT  & $89.6$ & $80.9$ & $87.6$ & $75.9$ & $80.5$ & $54.9$ & $78.2$ \\
    &&&&&&&&\\[-2.5mm]
    & \textbf{RobEn} & & & & & & &  \\
    & \concomptext + BERT & $80.6$ & $79.9$ & $84.2$ & $65.7$ & $73.3$& $52.7$ & $72.7$ \\ 
    & \agglomtext + BERT  & $83.1$ & $83.8$ & $85.0$ & $69.1$ & $76.6$ & $59.2$ & $76.1$ \\ 
    \hline
    &&&&&&&&\\[-2.5mm]
    \multirow{7}{*}{Attack} &  \textbf{Baselines} & & & & & & & \\
    & BERT & $8.7$ & $10.0$ & $17.4$ & $0.7$ & $0.7$ & $1.8$ & $6.6$ \\
    & Data Aug. + BERT & $17.1$ & $1.0$ & $27.6$ & $15.4$ & $10.7$ & $1.4$ & $12.2$ \\
    & Typo Corr. + BERT  & $53.2$ & $30.1$ & $52.0$ & $23.0$ & $32.3$ & $21.3$ & $35.3$ \\
    &&&&&&&&\\[-2.5mm]
    & \textbf{RobEn} & & & & & & &  \\
    & \concomptext + BERT & $80.3$ & $79.4$ & $82.7$ & $62.6$ & $71.5$ & $47.3$ & $70.6$ \\ 
    & \agglomtext + BERT  & $\bf 82.1$ & $\bf 82.8$ & $\bf 83.2$ & $\bf 65.3$ & $\bf 74.5$ & $\bf 52.7$ & $\bf 73.4$ \\ \hline
    &&&&&&&&\\[-2.5mm]
     \multirow{3}{*}{Robust} & \textbf{RobEn} & & & & & & &  \\
    & \concomptext + BERT & $80.1$ & $79.4$ & $\bf 82.2$ & $61.4$ & $70.5$ & $46.6$ & $70.0$ \\ 
    & \agglomtext + BERT  & $\bf 80.7$ & $\bf 80.9$ & $81.4$ & $\bf 62.8$ & $\bf 71.9$& $\bf 49.8$ & $\bf 71.3$ \\ 
    \hline
  \end{tabular}
  \caption{
  Standard, attack, and robust accuracy on six GLUE tasks against ED1 perturbations. For baseline models we only compute attack accuracy, an upper bound on robust accuracy,
  since robust accuracy cannot be tractably computed. 
  Using RobEn, we get robustness guarantees by computing robust accuracy, which we find outperforms a the typo corrector in \citep{pruthi2019misspellings} by \emph{at least} $36$ points.
  }
  \label{tab:main}
\end{table*}

Our main results are shown in \reftab{main}. 
We show all three baselines, as well as models using our instances of RobEn: \conncomp 
and \agglom.

Even against the heuristic attack, each baseline system suffers dramatic performance drops.
The system presented by \citet{pruthi2019misspellings}, Typo Corrector + BERT, only achieves $35.3\%$ attack accuracy, compared to its standard accuracy of $78.2\%$. BERT and Data Augmentation + BERT perform even worse.
Moreover, the number of perturbations the heuristic attack explores is a tiny fraction of our attack surface, so the robust accuracy of Typo Corrector + BERT, the quantity we'd like to measure, is likely far lower than the attack accuracy.

In contrast, simple instances of RobEn are much more robust. \agglom + BERT achieves average robust accuracy of $71.3\%$, $36$ points higher than the attack accuracy of Typo Corrector + BERT.
\agglom also further improves on \conncomp in terms of both robust accuracy (by $1.3$ points) and standard accuracy (by $2.8$ points).
\paragraph{Standard accuracy.} Like defenses against adversarial examples in other domains, using RobEn decreases standard accuracy \citep{madry2017towards,zhang2019theoretically,jia2019certified}.
Our agglomerative cluster encodings's standard accuracy is $10.1$ points lower then that of normally trained BERT. However, to the best of our knowledge, our standard accuracy is state-of-the-art for approaches that guarantee robustness. We attribute this improvement to RobEn's compatibility with any model.

\paragraph{Comparison to smaller attack surfaces.} 
We note that RobEn also outperform existing methods on their original, smaller attack surfaces. On SST-2, \citet{pruthi2019misspellings} achieves an accuracy of $75.0\%$ defending against a \emph{single} ED1 typo, which is $5.7$ points lower than $\agglom$'s robust accuracy against perturbations of all tokens: a superset of the original perturbation set. We discuss constrained adversaries further in Appendix \ref{apdx:constrained}. $\agglom$ also outperforms certified training: \citet{huang2019achieving}, which offers robustness guarantees to three character substitution typos (but not insertions or deletions), achieves a robust accuracy of $74.9\%$ on SST-2. 
Certified training requires strong assumptions on model architecture; 
even the robust accuracy of $\agglom$ outperforms the standard accuracy of the CNN used in \citet{huang2019achieving}. 
\begin{figure}[t]
    \centering
    \includegraphics[width = 0.8\linewidth]{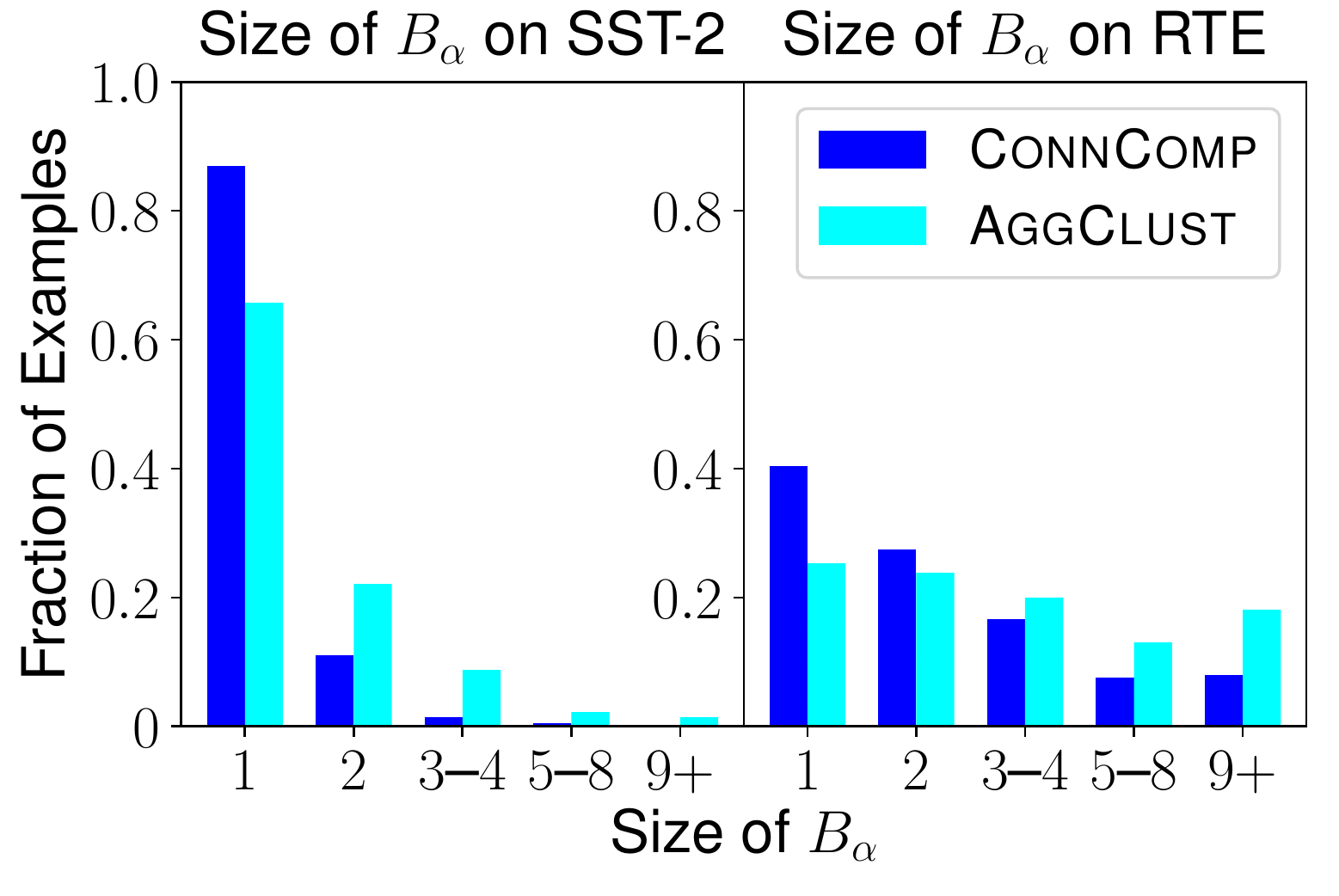}
    \caption{Histogram of $|\Brep(x)|$ for SST-2 and RTE. SST-2 has the highest percentage of inputs $x$ where $|\Brep(x)| = 1$, while RTE has the least. On both datasets, $|\Brep(x)| < 9$ for most $x$, and $|\Brep(x)| = 1$ on a plurality of inputs.
    }
    \label{fig:balpha_size}
\end{figure}
\subsection{Reusable encodings}
Each instance of RobEn achieves consistently high stability across our tasks, despite reusing a single function.
Figure \ref{fig:balpha_size} plots the distribution of $|\Brep(x)|$, across test examples in SST-2 and RTE, where $\Brep(x)$ is the set of encodings that are mapped to by some perturbation of $x$.
Over \agglom encodings, $|\Brep(x)| = 1$ for 
$25\%$ of examples in RTE and $66\%$ in SST-2, with the other four datasets falling between these extremes (see Appendix~\ref{apdx:balpha}).
As expected, these numbers are even higher for the connected component encodings.
Note that when $|\Brep(x)| = 1$, every perturbation of $x$ maps to the same encoding. When $|\Brep(x)|$ is small, robust accuracy can be computed quickly.
\subsection{Agglomerative Clustering Tradeoff} 
\label{apdx:aggtradeoff}
\begin{figure}
    \centering
    \includegraphics[width = 0.8\linewidth]{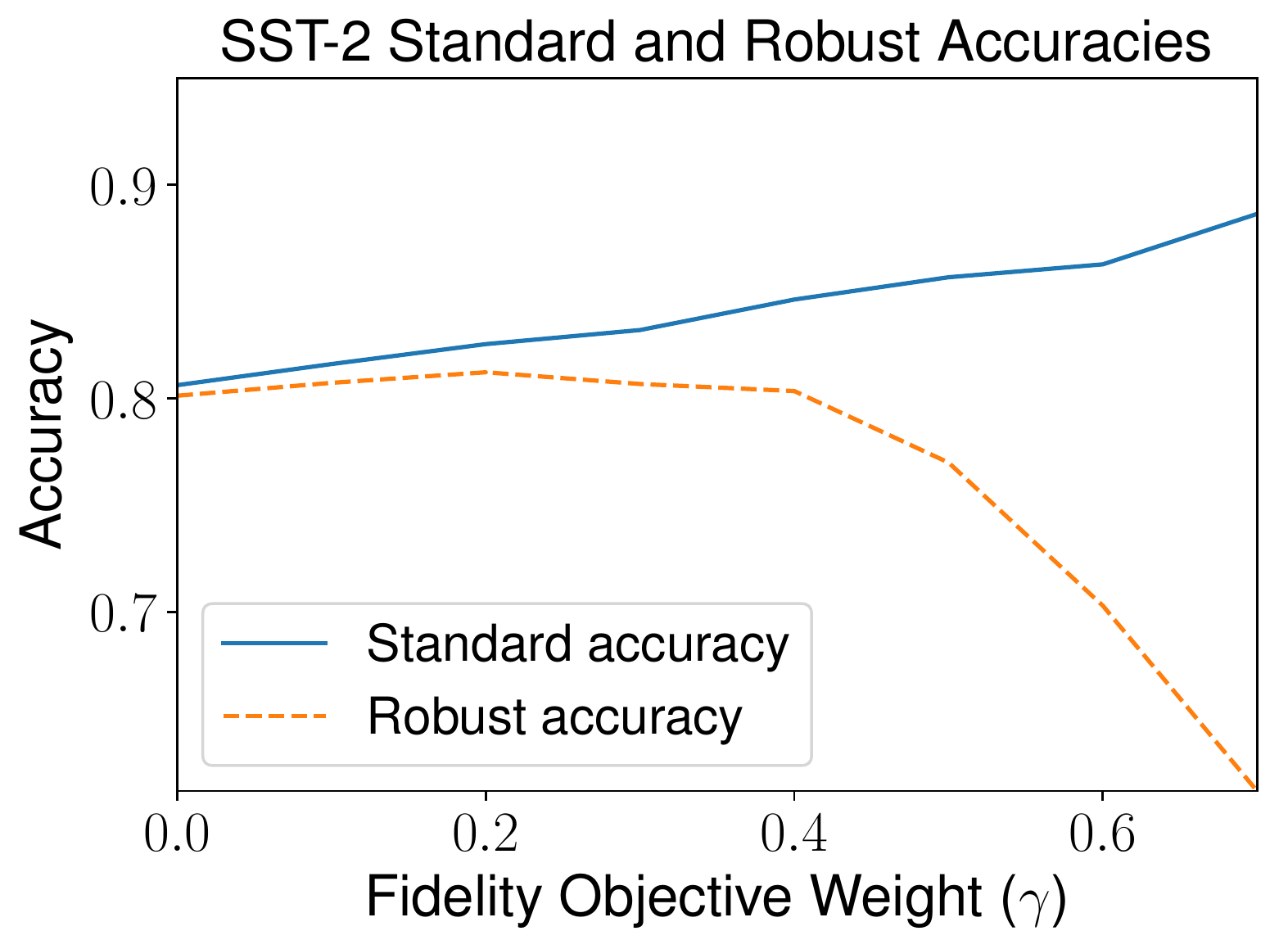}
    \caption{Standard and robust accuracies on SST-2 with \agglom using different values of $\gamma$. While the gap between standard and robust accuracy increases monotonically, robust accuracy increases before decreasing.}
    \label{fig:tradeoff}
\end{figure}
In \reffig{tradeoff}, we plot standard and robust accuracy on SST-2 for \agglom encodings, using different values of $\gamma$.
Recall that $\gamma = 0$ maximizes stability (\conncomp), and $\gamma=1$ maximizes fidelity.  
At $\gamma=0$, the gap between standard and robust accuracy, due to out-of-vocabulary tokens, 
is negligible. As $\gamma$ increases, both standard accuracy and the gap between standard and robust accuracy increase. As a result, robust accuracy first increases, then decreases.

\subsection{Internal permutation attacks}
\label{sec:intprm}
RobEn can also be used to defend against the internal perturbations described in Section \ref{sec:experiments_setup}. 
For normally trained BERT, a heuristic beam search attack using internal permutations reduces average accuracy from $86.2\%$ to $15.7\%$ across our six tasks. 
Using \conncomp with the internal permutation attack surface, we achieve robust accuracy of $81.4\%$. See Appendix~\ref{apdx:intprm} for further details.
\section{Discussion}
\paragraph{Additional related work.} In this work, we introduce RobEn, a framework to construct systems that are robust to adversarial perturbations. We then use RobEn to achieve state-of-the-art robust accuracy when defending against adversarial typos. Besides typos, other perturbations can also be applied to text.
Prior attacks consider semantic operations, such as replacing a word with a synonym \citep{alzantot2018adversarial, ribeiro2018sears}. Our framework extends easily to these perturbations.
Other attack surfaces involving insertion of sentences \citep{jia2017adversarial} or syntactic rearrangements \citep{iyyer2018adversarial} are harder to pair with RobEn, and are interesting directions for future work.

Other defenses are based on various forms of preprocessing. \citet{gong2019context} apply a spell-corrector to correct typos chosen to create ambiguity as to the original word, but these typos are not adversarially chosen to fool a model.
\citet{edizel2019misspelling} attempt to learn typo-resistant word embeddings, 
but focus on common typos, rather than worst-case typos.
In computer vision, \citet{chen2019towards} discretizes pixels to compute exact robust accuracy on MNIST, but their approach generalizes poorly to other tasks like CIFAR-10. \citet{garg2018robust} generate functions that map to robust features, while enforcing variation in outputs.

\paragraph{Incorporating context.} Our token-level robust encodings lead to strong performance, despite ignoring useful contextual information. 
Using context is not fundamentally at odds with the idea of robust encodings, and making contextual encodings stable is an interesting technical challenge and a promising direction for future work.

In principle, an oracle that maps every word with a typo to the correct unperturbed word seems to have higher fidelity than our encodings, without compromising stability.
However, existing typo correctors are far from perfect, and a choosing an incorrect unperturbed word from a perturbed input leads to errors in predictions of the downstream model.
This mandates an intractable search over all perturbations to compute the robust accuracy.

\paragraph{Task-agnosticity.}
Many recent advances in NLP have been fueled by the rise of task-agnostic representations, such as BERT, that facilitate the creation of accurate models for many tasks. 
Robustness to typos should similarly be achieved in a task-agnostic manner, as it is a shared goal across many NLP tasks.
Our work shows that even simple robust encodings generalize across tasks and are more robust than existing defenses.
We hope our work inspires new task-agnostic robust encodings that lead to more robust and more accurate models.

\subsubsection*{Acknowledgments}
This work was supported by NSF Award Grant no. 1805310 and the DARPA ASED program under FA8650-18-2-7882. A.R. is supported by a Google PhD
Fellowship and the Open Philanthropy Project AI Fellowship. We thank Pang Wei Koh, Reid Pryzant, Ethan Chi, Daniel Kang, and the
anonymous reviewers for their helpful comments.
\subsubsection*{Reproducibility}
All code, data, and experiments are available on
CodaLab at {\large\href{https://bit.ly/2VSZI2e}{https://bit.ly/2VSZI2e}}. 

\bibliography{main.bib}
\bibliographystyle{acl_natbib.bst}

\appendix
\section{Appendix}

	\subsection{Aggloemrative clustering}
	\label{apdx:aggalgo}
	
Recall that any $\phitokbar$ induces a clustering of $V$,
where each cluster contains a set of words mapped by $\phitokbar$ to the same encoded token.
We use an agglomerative clustering algorithm to approximately minimize $\Phi$.
We initialize $\phitokbar$ by setting $\phitokbar(w) = w$ for each $w \in V$,
which corresponds to placing each word in its own cluster.
We then examine each pair of clusters $C_i, C_j$ such that 
there exists an edge between a node in $C_i$ and a node in $C_j$, 
in the graph from \refsec{tokenexample}. 
For each such pair,
we compute the value of $\Phi$ if $C_i$ and $C_j$ were replaced by $C_i \cup C_j$. 
If no merge operation causes $\Phi$ to decrease, we return the current $\phitokbar$.
Otherwise, we merge the pair that leads to the greatest reduction in $\Phi$, and repeat.
To merge two clusters $C_i$ and $C_j$,
we first compute a new encoded token $r$ as the $w \in C_i \cup C_j$ with largest $\rho(w)$.
We then set $\phitokbar(w) = r$ for all $w \in C_i \cup C_j$. Our algorithm thus works as follows
	\begin{algorithm}[H]
		\caption{Objective-minimizing agglomerative clustering}\label{alg:aggclu}
		\begin{algorithmic}[1]
		\STATE {$C \gets V$}
		\FOR{$i$ in range($|V|$)}
		    \STATE {$C_\text{next} \gets \text{Get Best Combination}(C)$}
		    \IF{$C = C_\text{next}$}
		        \STATE {\textbf{return} C}
		    \ENDIF
		    \STATE {$C \gets C_\text{next}$}
		\ENDFOR
	    \STATE {\textbf{return}} C
	    \end{algorithmic}
	\end{algorithm}
	Now, we simply have to define the procedure we use to get the best combination. 
	\begin{algorithm}[H]
		\caption{$\text{Get Best Combination}(C)$}\label{alg:aggclu}
		\begin{algorithmic}[1]
		\floatname{algorithm}{Get Best Combination}
		    \STATE {$C_\text{opt} \gets C$}
		    \STATE {$\Phi_\text{opt} \gets \Phi(C)$}
		    \FOR{$(C_i, C_j) \in \text{Adjacent Pairs}(C)$}
		        \STATE {$C_\text{comb} \gets C_i \cup C_j$}
		        \STATE {$C_{\text{new}} \gets C \cup C_\text{comb} \setminus \{C_i, C_j\}$} \COMMENT{New clusters}
		        \STATE {$\Phi_\text{new} \gets \Phi(C_{\text{new}})$}
		        \IF {$\Phi_{new} < \Phi_{opt}$}
		            \STATE {$\Phi_{opt} \gets \Phi_{new}$}
		            \STATE {$C_\text{opt} \gets C_{\text{new}}$}
		        \ENDIF
		    \ENDFOR
		    \STATE { \textbf{return} $C_\text{opt}$ } 
		\end{algorithmic}
	\end{algorithm}
	Recall our graph $G = (G, E)$ used to define the connected component clusters. We say two clusters $C_i$ and $C_j$ are \emph{adjacent}, and thus returned by $\text{Adjacent Pairs}$, if there exists a $v_i \in C_i$ and a $v_j \in C_j$ such that $(v_i, v_j) \in G_E$. The runtime of our algorithm is $O(N^2E)$ since at each of a possible $N$ total iterations, we compute the objective for one of at most $E$ pairs of clusters. Computation of the objective can be reframed as computing the difference between $\Phi$ and $\Phi_\text{new}$, where the latter is computed using new clusters, which can be done in $O(N)$ time.

\subsection{Attacks}
\label{apdx:attacks}
We use two heuristic attacks to compute an upper bound for robust accuracy: one for ED1 perturbations and one for internal permutations. Each heuristic attack is a beam search, with beam width 5. However, because $|B(x_i)|$ is very large for many tokens $x_i$, even the beam search is intractable. Instead, we run a beam search where the allowable perturbations are $B^\prime(x_i) \subseteq B(x_i)$, where $|B^\prime(x_i)| << B(x_i)$ for sufficiently long $x_i$. 
For our ED1 attack, we define $B^\prime(x_i)$ to be four randomly sampled perturbations from $B(x_i)$ when the length of $x_i$ is less than five, and all deletions when $x_i$ is greater than five. Thus, the number of perturbations of each word is bounded above by $\min\{4, \text{len}(x_i) - 2\}$. For our internal permutations, $B^\prime(x_i)$ is obtained by sampling five permutations at random.

\subsection{Datasets}
\label{sec:datasets}
We use six out of the nine tasks from GLUE: SST, MRPC, QQP, MNLI, QNLI, and RTE, all of which are classification tasks measured by accuracy. The Stanford Sentiment Treebank (SST-2) \citep{socher2013recursive} contains movie reviews that are classified as positive and negative. The Microsoft Research Paraphrase Corpus (MRPC) \citep{dolan2005mrpc} and the Quora Question Pairs dataset\footnote{data.quora.com/First-Quora-Dataset-Release-Question-Pairs} contain pairs of input which are classified as semantically equivalent or not; QQP contains question pairs from Quora, while MRPC contains pairs from online news sources. MNLI, and RTE are entailment tasks, where the goal is to predict whether or not a premise sentence entails a hypothesis \citep{williams2018broad}. MNLI gathers premise sentences from ten different sources, while RTE gathers premises from entailment challenges. QNLI gives pairs of sentences and questions extracted from the Stanford Question Answering Dataset \citep{rajpurkar2016squad}, and the task is to predict whether or not the answer to the question is in the sentence.

We use the GLUE splits for the six datasets and evaluate on test labels when available (SST-2, MRPC), and otherwise the publicly released development labels. We tune hyperparameters by training on 80\% of the original train set and using the remaining 20\% as a validation set. We then retrain using the chosen hyperparameters on the full training set.

\subsection{Experimental details}
\label{apdx:hparams}
For our methods using transformers, we start with the pretrained uncased BERT \citep{devlin2019bert}, using the same hyperparameters as the pytorch-transformers repo.\footnote{\url{https://github.com/huggingface/pytorch-transformers}}. In particular, we use the base uncased version of BERT. We use a batch size of 8, and learning rate $2e{-5}$. For examples where $|\Brep(x)| > 10000$, we assume the prediction is not robust to make computation tractible. Each typo corrector uses the defaults for training from\footnote{\url{https://github.com/danishpruthi/Adversarial-Misspellings}}; it is trained on a specific task using perturbations of the training data as input and the true sentence (up to OOV) as output. The vocabulary size of the typo correctors is 10000 including the unknown token, as in \citep{pruthi2019misspellings}. The typo corrector is chosen based on word-error rate on the validation set. 

\subsection{Constrained adversaries}
\label{apdx:constrained}
Using RobEn, since we can tractably compute robust accuracy, it is easy to additionally consider adversaries that cannot perturb every input token. We may assume that an attacker has a budget of $b \le L$ words that they may perturb as in \citep{pruthi2019misspellings}.
Exiting methods for certification \citep{jia2019certified, huang2019achieving} require attack to be factorized over tokens, and cannot give tighter guarantees in the budget-constrained case compared to the unconstrained setting explored in previous sections.
However, our method lets us easily compute robust accuracy exactly in this situation:
we just enumerate the possible perturbations that satisfy the budget constraint, and query the model.
\begin{figure}
    \centering
    \includegraphics[width=0.95\linewidth]{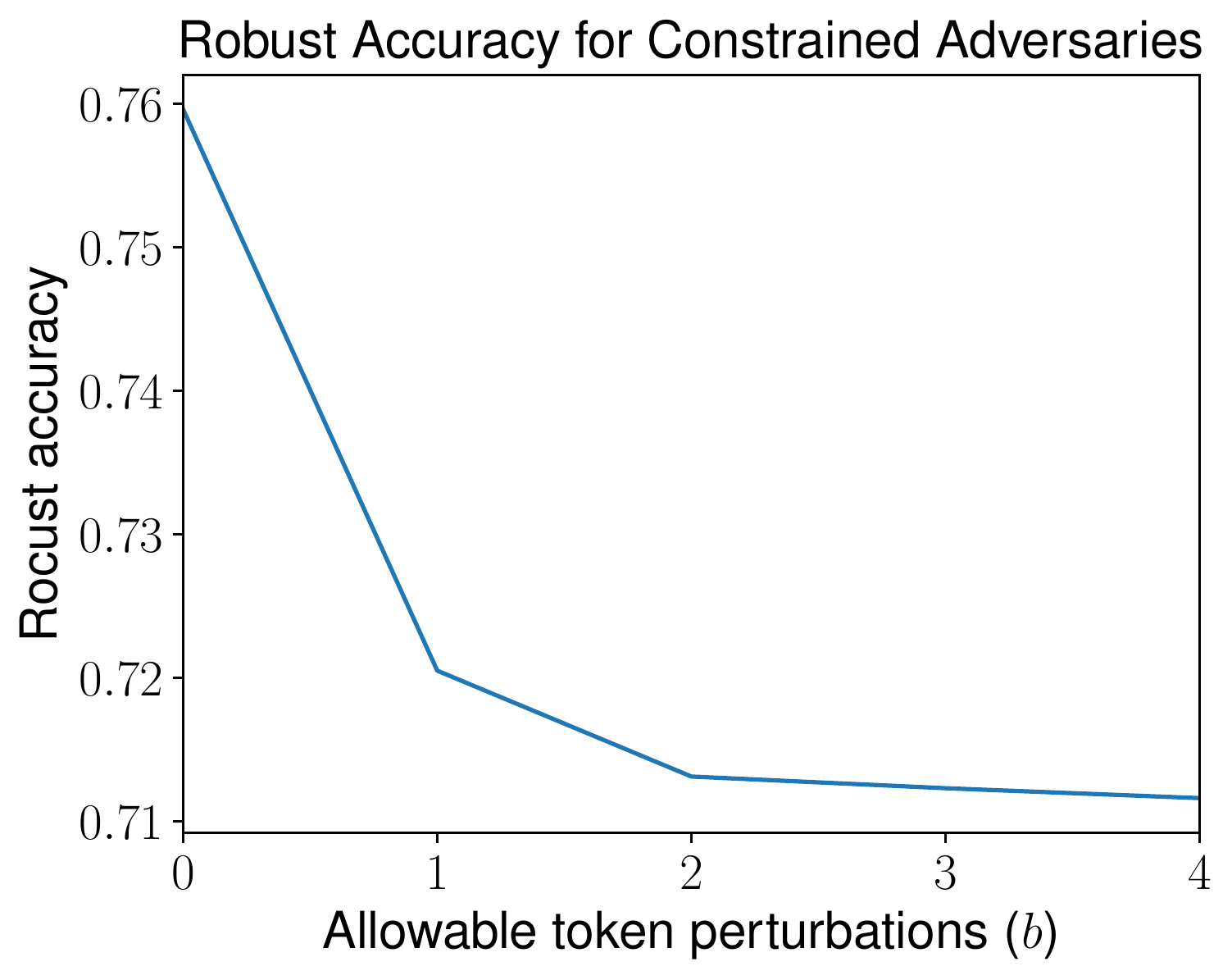}
    \caption{Robust accuracy averaged across all tasks based on different adversarial budgets $b$. $b = 0$ corresponds to clean performance, and robust performance is reached at $b = 4$}
    \label{fig:budget}
\end{figure}

\reffig{budget} plots average robust accuracy across the six tasks using \agglom\text{ } as a function of $b$. Note that $b=0$ is simply standard accuracy. 
Interestingly, for each dataset there is an attack only perturbing 4 tokens with attack accuracy equal to robust accuracy.

\subsection{Number of representations}
\label{apdx:balpha}
\begin{figure}
\centering   
\subfigure[Size of $\Brep$ over MRPC and QQP]{\label{fig:mrpc_qqp_balpha_size}\includegraphics[width=68mm]{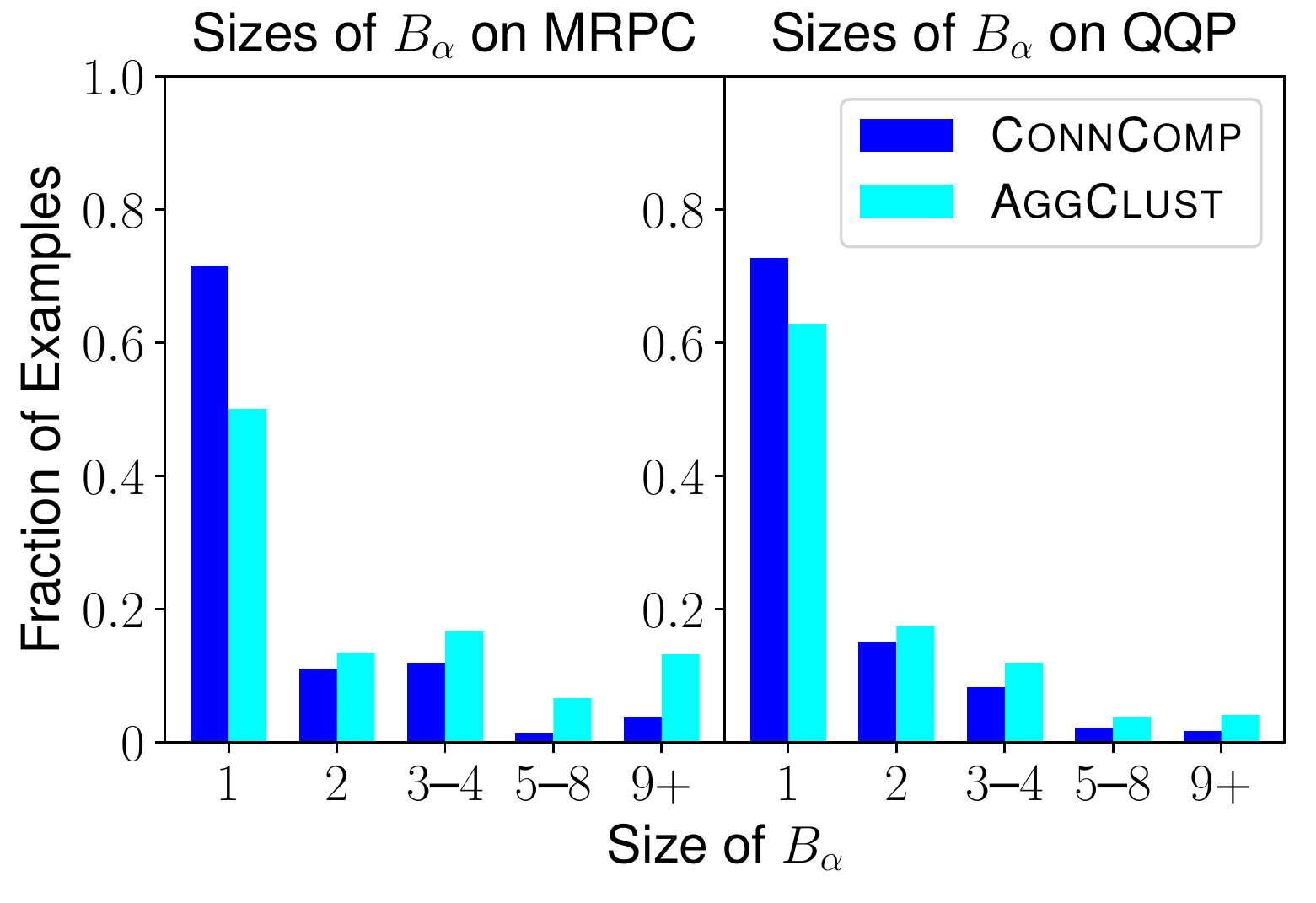}}
\subfigure[Size of $\Brep$ over MNLI and QNLI]{\label{fig:mnli_qnli_balpha_size}\includegraphics[width=68mm]{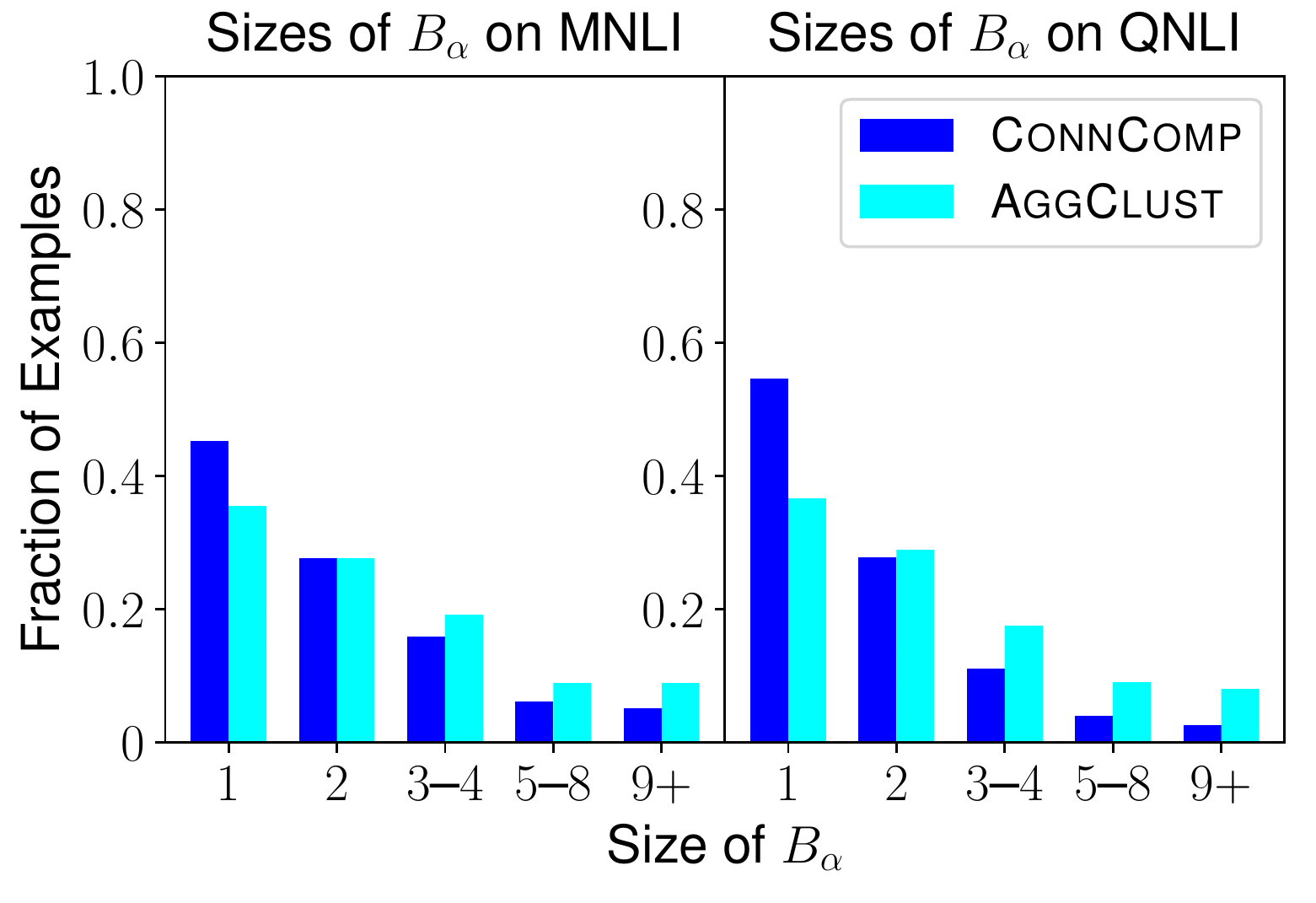}}
\caption{Histograms showing sizes of $\Brep$ for MRPC, QQP, MNLI, and QNLI.}
\end{figure}
We include here histograms for the datasets we did not cover in the main body. The histograms for MRPC and QQP are shown in Figure \ref{fig:mrpc_qqp_balpha_size}, while the histograms for MNLI and QNLI are shown in Figure \ref{fig:mnli_qnli_balpha_size}. The fraction of $x$ such that $|\Brep(x)| = 1$ for each dataset and each set of encodings is provided in Table \ref{tab:numone}.
\begin{table*}[t]
  \small
  \centering
  \begin{tabular}{|l|rrrrrrr|}
    \hline
    &&&&&&&\\[-2.5mm]
    Encodings & SST-2 & MRPC & QQP & MNLI & QNLI & RTE & Avg \\\hline
    \concomptext & $86.9$ & $71.6$ & $72.7$ & $45.3$ & $54.6$ & $40.4$ & $61.9$ \\
    \agglomtext & $65.6$ & $50.0$ & $62.7$ & $35.4$ & $36.6$ & $25.2$ & $45.9$ \\\hline
  \end{tabular}
  \caption{Percentage of test examples with $|\Brep(x)| = 1$ for each dataset.}
  \label{tab:numone}
\end{table*}

\subsection{Internal Permutation Results}
\label{apdx:intprm}
We consider the \emph{internal permutation} attack surface, where interior characters in a word can be permuted, assuming the first and last characters are fixed. For example, \nl{perturbation} can be permuted to \nl{peabreuottin} but not \nl{repturbation}. Normally, context helps humans resolve these typos. Interestingly, for internal permutations it is impossible for an adversary to change the cluster assignment of both in-vocab and out of vocab tokens since a cluster can be uniquely represented by the first character, a sorted version of the internal characters, and the last character. Therefore, using \conncomp\text{ }encodings, robust, attack, and standard accuracy are all equal. We use the attack described in \ref{apdx:attacks} to attack the clean model. The results are in Table \ref{tab:intprm}.
\begin{table*}[t]
  \small
  \centering
  \begin{tabular}{|l|l|rrrrrrr|}
    \hline
    Accuracy & \multicolumn{1}{c|}{System} 
    & \multicolumn{1}{c}{SST-2}
    & \multicolumn{1}{c}{MRPC}
    & \multicolumn{1}{c}{QQP}
    & \multicolumn{1}{c}{MNLI}
    & \multicolumn{1}{c}{QNLI}
    & \multicolumn{1}{c}{RTE}
    & \multicolumn{1}{c|}{Avg} \\ \hline
    &&&&&&&&\\[-2.5mm]
    \multirow{2}{*}{Standard} & BERT & $93.8$ & $87.7$ & $91.2$ & $84.3$ & $88.9$ & $71.1$ & $86.2$ \\
    & \concomptext + BERT & $93.2$ & $87.7$ & $86.9$ & $75.9$ & $83.4$ & $61.4$ & $81.4$ \\\hline
    \multirow{2}{*}{Attack} & BERT & $28.1$ & $15.9$ & $33.0$ & $4.9$ & $6.2$ & $5.8$ & $15.7$ \\
    & \concomptext + BERT & $93.2$ & $87.7$ & $86.9$ & $75.9$ & $83.4$ & $61.4$ & $81.4$ \\\hline
    &&&&&&&&\\[-2.5mm]
    Robust & \concomptext + BERT & $93.2$ & $87.7$ & $86.9$ & $75.9$ & $83.4$ & $61.4$ & $81.4$ \\
    \hline
  \end{tabular}
  \caption{Results from internal permutation attacks. Internal permutation attacks bring the average performance for BERT across the six listed tasks from $86.2$ to $15.7$. Our \conncomp\text{ } encodings, generated using the internal permutation attack surface, achieve a robust accuracy of $81.4$, which is only $4.8$ points below standard accuracy.}
  \label{tab:intprm}
\end{table*}

\end{document}